%% file: PaperForReview.tex
\documentclass[10pt,twocolumn,letterpaper]{article}
 \usepackage[T1]{fontenc}     

\usepackage[pagenumbers]{wacv} 

\usepackage{graphicx}
\usepackage{amsmath}
\usepackage{amssymb}
\usepackage{booktabs}

\usepackage[pagebackref,breaklinks,colorlinks]{hyperref}

\input{preamble}
\usepackage[capitalize]{cleveref}
\crefname{section}{Sec.}{Secs.}
\Crefname{section}{Section}{Sections}
\Crefname{table}{Table}{Tables}
\crefname{table}{Tab.}{Tabs.}

\def\wacvPaperID{3411} 
\def\confName{WACV}
\def\confYear{2025}

\begin{document}

\title{\textsc{CAMBench-QR}: A Structure-Aware Benchmark for Post-Hoc \\Explanations with QR Understanding}
\author{
Ritabrata Chakraborty$^{1}$ \quad Avijit Dasgupta$^{2}$ \quad Sandeep Chaurasia$^{1}$ \\
$^{1}$ Manipal University Jaipur \qquad
$^{2}$ IIIT Hyderabad
}

\maketitle

\begin{abstract}
Visual explanations are often plausible but not structurally faithful. We introduce CAMBench-QR, a structure-aware benchmark that leverages the canonical geometry of QR codes (finder patterns, timing lines, module grid) to test whether CAM methods place saliency on requisite substructures while avoiding background. CAMBench-QR synthesizes QR/non-QR data with exact masks and controlled distortions, and reports structure-aware metrics (Finder/Timing Mass Ratios, Background Leakage, coverage AUCs, Distance-to-Structure) alongside causal occlusion, insertion/deletion faithfulness, robustness, and latency. We benchmark representative, efficient CAMs (LayerCAM, EigenGrad-CAM, XGrad-CAM) under two practical regimes of zero-shot and last-block fine-tuning. The benchmark, metrics, and training recipes provide a simple, reproducible yardstick for structure-aware evaluation of visual explanations. Hence we propose that CAMBENCH-QR can be used as a litmus test of whether visual explanations are truly structure-aware.

\end{abstract}


\section{Introduction}
\label{sec:intro}
\epigraph{``Are you watching closely?''}{Alfred Borden, \textit{The Prestige}}

\begin{figure}[t] \centering \includegraphics[width=\linewidth]{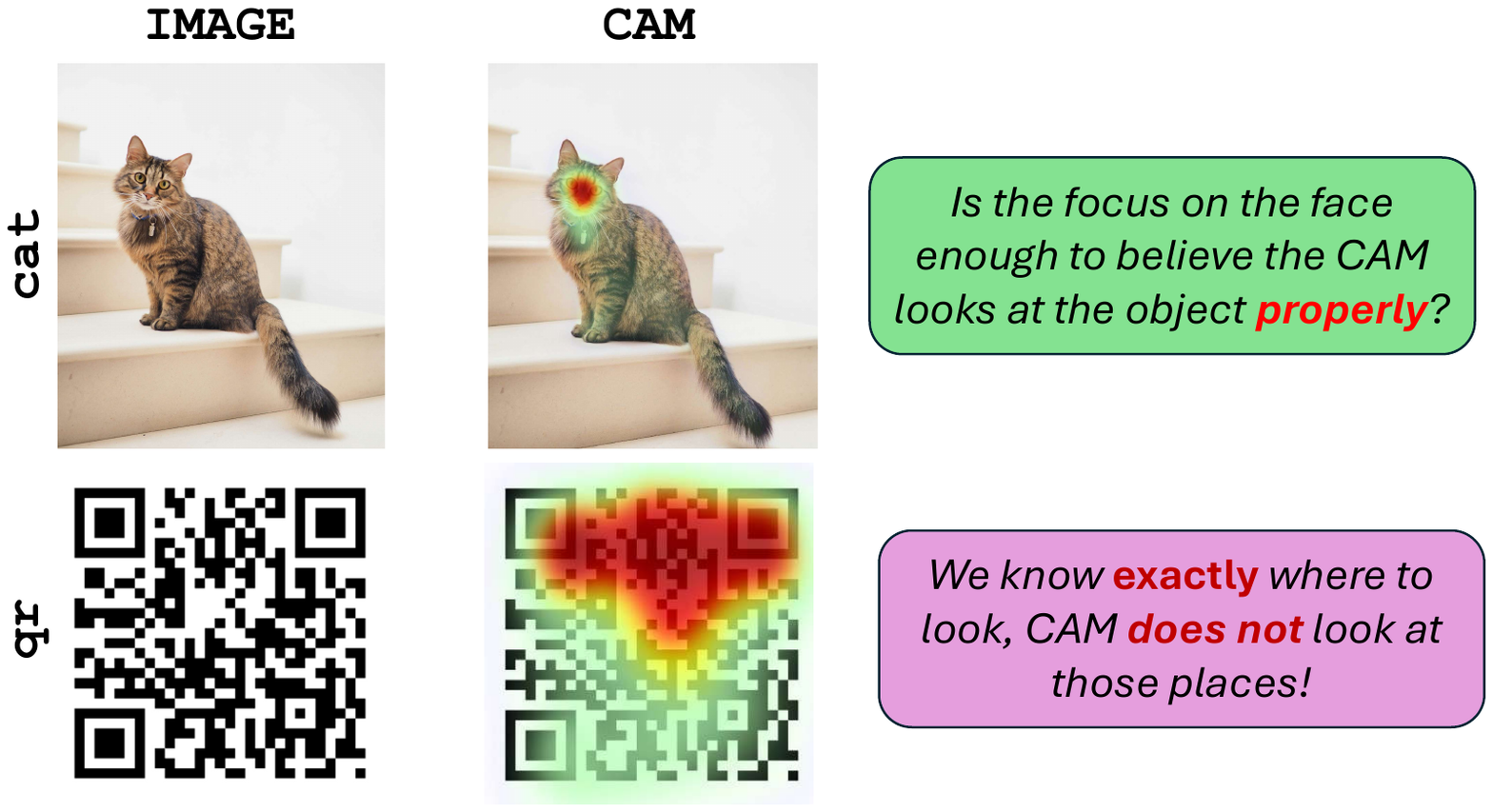} \caption{\textbf{Plausible vs.\ structural explanations.} (Top) On natural images, CAMs may look convincing without ground truth for “where to look.” (Bottom) For QR codes, canonical parts are known, allowing objective, structure-aware evaluation.} \label{fig:fig1} \end{figure}

The question is fitting for explainable vision: saliency maps can be compelling misdirection. On natural images, visually plausible heatmaps often look right without being \emph{correct}. When the target class exhibits \emph{rigid geometry} and \emph{canonical parts}, explanations should be \emph{structurally faithful}: attribution ought to concentrate on the parts that constitute the concept and avoid incidental texture that merely correlates with the label. Figure~\ref{fig:fig1} contrasts these settings: a cat admits many aesthetically reasonable explanations, whereas a QR code has non-negotiable components—three finder patterns, timing lines, and a module grid—so the “where to look” is objectively knowable.
Structural faithfulness matters because CAM explanations inform auditing, debugging, and deployment decisions. If evaluation rewards plausible masks instead of \emph{part-level alignment}, models can appear trustworthy while relying on shortcut evidence (borders, backgrounds, repetitive textures). This risk is acute in high-frequency or layout-driven domains—codes, scientific imagery, documents—where texture and geometry are easily confounded: a classifier may leverage backgrounds, and a CAM may faithfully reflect that shortcut. A structure-aware perspective reframes the question from \textit{does the map look reasonable?} to \textit{does the evidence land on the parts that make the concept work—and stay off everything else?}

QR codes offer an unusual opportunity to study this precisely. Their geometry is canonical and parameterizable, enabling synthesis of QR/non-QR data with exact part masks and controlled distortions. This permits measurements that standard pixel-overlap proxies cannot cleanly capture: how much saliency mass falls on finder and timing structures versus background; how the confident core covers required parts across thresholds; how far spurious mass drifts from structure; and whether structural mass predicts causal influence when parts are occluded or gradually inserted/deleted. Because the generation process is controlled, we can apply rotation, perspective, blur, JPEG compression, low light, and occlusion while preserving ground-truth structure.

To interpret results beyond any one method, we organize CAMs along three orthogonal axes (Fig.~\ref{fig:cam_axes}): how pixel importance is computed (gradients, activations, perturbations), how feature maps are aggregated (global pooling, layer-wise fusion, subspace denoising), and computational profile (single-pass lightweight versus multi-mask/ablation). We further separate the \emph{explainer rule} from the \emph{feature basis} it aggregates: keeping the rule fixed while varying trainable depth changes the basis (last block versus deeper) without altering inference-time cost. This factorization lets us examine zero-shot use (frozen backbone + linear head), last-block fine-tuning, and a simple leakage-minimizing penalty in a controlled way.

\begin{tcolorbox}[
  enhanced,
  breakable,
  colback=green!6!white,
  colframe=green!50!black,
  title={Research Questions},
  fonttitle=\bfseries,
  boxrule=0.6pt,
  left=1mm,right=1mm,top=1mm,bottom=1mm
]
\begin{description}[leftmargin=0pt,labelsep=0.5em]
  \item[\textbf{RQ1}] Do structure-aware metrics---mass on finder/timing modules, background leakage, coverage AUCs, and distance-to-structure---agree with causal tests such as part-occlusion correlations and insertion/deletion curves?
  \item[\textbf{RQ2}] How does trainable depth (head-only $\rightarrow$ last block $\rightarrow$ deeper) reshape attribution structure when the explainer rule is held fixed?
  \item[\textbf{RQ3}] Can a light penalty suppress off-structure mass without collapsing in-QR coverage?
  \item[\textbf{RQ4}] How stable are structure-aware explanations under common distortions?
\end{description}
\end{tcolorbox}

\noindent\textbf{Contributions.} Our contributions can be summarized as follows:
\begin{itemize}
    \item \textbf{A structure-aware dataset and protocol.} Synthetic QR/non-QR corpora with exact finder, timing, and box masks plus controlled distortions, enabling geometry-aware evaluation beyond pixel-overlap proxies.
    \item \textbf{Metrics that target parts and avoid shortcuts.} Mass-on-structure ratios (FMR/TMR), background leakage (BL), structure-coverage AUCs, and normalized distance-to-structure (DtS), complemented by targeted part-occlusion correlations, insertion/deletion faithfulness, and latency for deployability.
    \item \textbf{A factorized study across methods and regimes.} A principled axes-based comparison of representative, efficient CAM families under two practical regimes—zero-shot and last-block fine-tuning—with an optional leakage-minimizing penalty; experiments span ResNet-50~\cite{he2016deep} and ConvNeXt-B~\cite{liu2022convnet}.
    \item \textbf{Simple training recipes that improve structure without architectural changes.} Short schedules that adapt only the last residual stage to re-orient high-level features toward canonical parts, and a lightweight penalty that suppresses off-structure mass, both with negligible inference-time overhead.
\end{itemize}


The rest of the paper is structured as follows. Section \ref{sec:related_works} discusses related literature to CAM models, explanation evaluations and structure aware studies.
Section \ref{sec:method} discusses the pipeline of CAMBENCH-QR, taxonomy of the axes of our study and how we use the proposed metrics for structural evaluation. This is consolidated by Section \ref{sec:results} which shows quantitative results for the benchmark. We discuss findings and takeaways in Section \ref{sec:discussion} and conclude the paper in Section \ref{sec: conclusion}.

\section{Related Works}
\label{sec:related_works}

\paragraph{CAM and its modern variants.}
Class Activation Mapping (CAM) demonstrated that inserting global average pooling allowed models trained with image-level labels to still reveal spatially discriminative regions, sparking a large family of class-discriminative heatmaps \cite{Zhou_2016_CVPR}. Grad-CAM removed the architectural constraint of CAM by using gradients to generalize to arbitrary CNNs and tasks \cite{Selvaraju_2017_ICCV}. Ideas consequent to this centered around sharpening or stabilizing maps. For instance, Grad-CAM++ improved multi-instance localization and weighting \cite{8354201}, while Smooth Grad-CAM++ smoothened activations for crisper maps~\cite{omeiza2019smooth}. A parallel, gradient-free branch weights features via forward signals: Score-CAM~\cite{wang2020score}, with refinements SS-CAM~\cite{wang2020ss} and IS-CAM~\cite{naidu2020cam}. Ablation-CAM estimates causal importance by ablating feature maps~\cite{9093360}. To reduce query/runtime cost, \textit{Group-CAM} aggregates and perturbs grouped activations~\cite{zhang2021group}; \textit{ReciproCAM} proposes a lightweight, gradient-free variant via spatial perturbations~\cite{byun2024reciprocam}. Axiomatic work formalizes desiderata such as Sensitivity/Conservation in \textit{XGrad-CAM}~\cite{fu2020axiom}. Finally, to capture detail across scales, \textit{Layer-CAM} aggregates evidence from multiple layers~\cite{jiang2021layercam}, while \textit{Eigen-CAM} replaces gradients with a PCA projection of activations for object-focused maps \cite{muhammad2020eigen}.

\paragraph{CAM-style explanations for Transformers.}
CAM-style relevance propagation has been adapted to attention architectures. Chefer \textit{et al.} derive class-specific relevancy for pure self-attention Transformers and extend to encoder--decoder and bimodal settings, enabling explanations across ViT/DETR/VQA-style models~\cite{Chefer_2021_CVPR,chefer2021generic}.

\paragraph{Formal evaluation of explanations.}
Early quantitative proxies emphasized localization. The Pointing Game, introduced with Excitation Backprop counts hits when the max-saliency lands inside the target region~\cite{zhang2018top}. Sanity checks revealed methods that are insensitive to model parameters or data, emphasizing validity over aesthetics~\cite{adebayo2018sanity}. Causal faithfulness is probed by \emph{ROAR} (remove top regions and retrain; larger accuracy drops imply more faithful attributions)~\cite{hooker2019benchmark}, and by black-box \emph{Insertion/Deletion} curves via \emph{RISE}~\cite{petsiuk2018rise}. Theoretically grounded criteria like \emph{Infidelity} and \emph{Sensitivity-$n$} formalize agreement with output changes and robustness to perturbations~\cite{yeh2019fidelity}. Across these axes, no single CAM/gradient method dominates, motivating multi-criteria evaluation rather than single-number scoring.

\paragraph{Structure-aware and part-level benchmarks.}
Recent work moves beyond pixel masks to test whether explanations capture \emph{structured} patterns. Scientific Reports introduced a ground-truth saliency comparison and an $m_{\mathrm{GT}}$ mass-overlap metric to quantify energy on annotated regions~\cite{szczepankiewicz2023ground}. \emph{FunnyBirds} provides a synthetic dataset with part-level interventions, mapping diverse explanations to a common space of part importances and enabling compositional analysis~\cite{Hesse_2023_ICCV}. A 2024 fidelity study reports a trade-off: backprop-style methods often yield higher fidelity than CAM variants but can produce noisier maps, advocating evaluation that balances localization, causality, robustness, and structure alignment~\cite{miro2024assessing}.


\begin{figure*}[t]
  \centering
  \includegraphics[width=\textwidth]{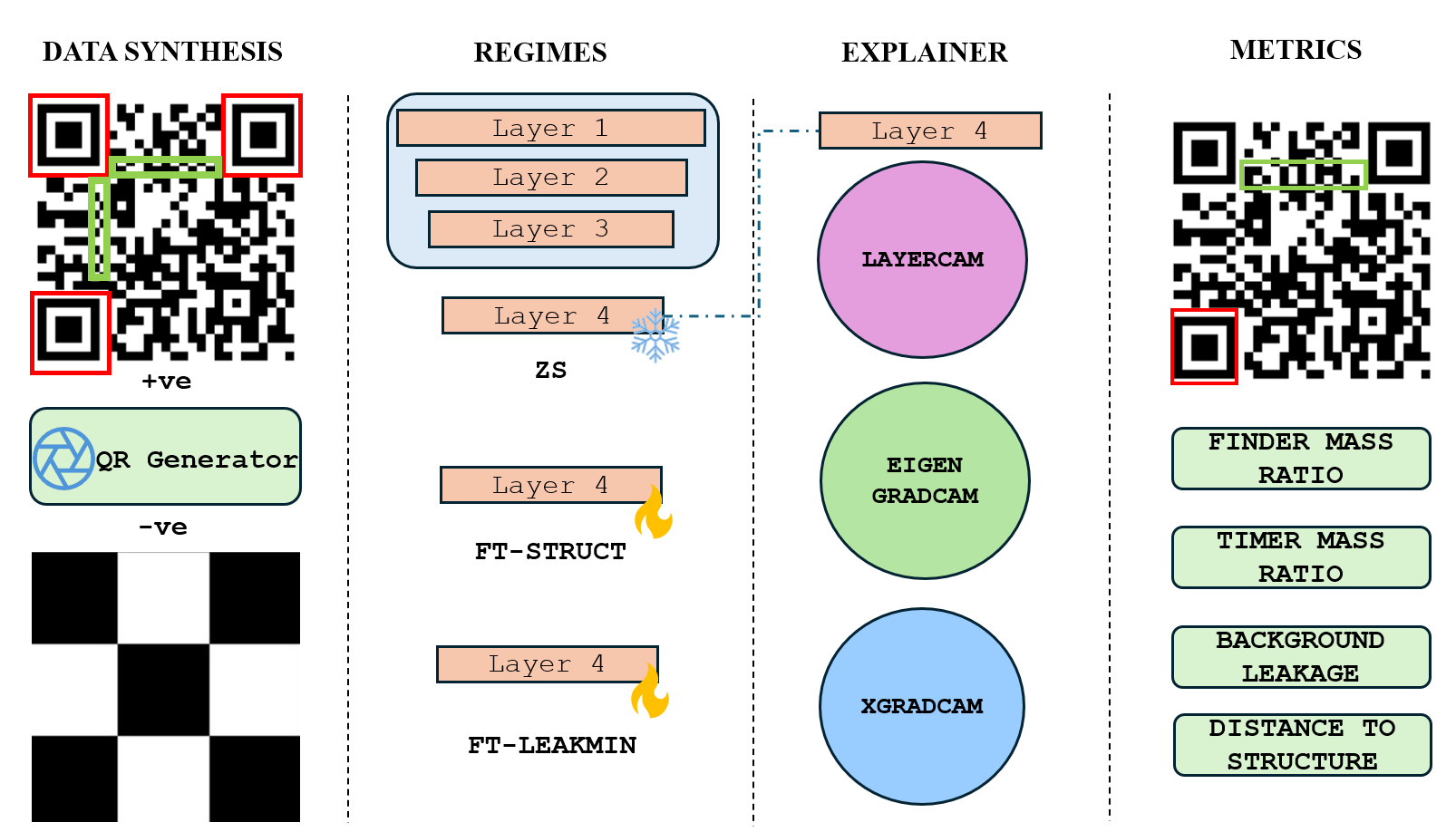}
  \vspace{-0.5cm}
  \caption{\textbf{CAMBench-QR pipeline.} Left→right: \emph{Data}, \emph{Regimes}, \emph{Explainer}, \emph{Metrics}. 
\textbf{Data:} synthesize QR positives with exact finder/timing masks and hard negatives (e.g., checkerboards); red boxes = finders, green band = timing; masks persist under distortions. 
\textbf{Regimes:} ZS (backbone frozen), FT-Struct (layer4 only), FT-LeakMin (layer4 + leakage penalty). 
\textbf{Explainer:} LayerCAM, EigenGrad-CAM, XGrad-CAM applied to the same model, operating on layer4 features. 
\textbf{Metrics:} FMR/TMR (mass on parts), BL (off-QR), DtS (drift); aggregated over images and distortion sweeps for fair, structure-aware comparison.}
  \label{fig:pipeline}
\end{figure*}
\section{CAMBENCH-QR}
\label{sec:method}

\subsection{CAM Families Along Three Axes}
\begin{figure}[H]
    \centering
    \includegraphics[width=\linewidth]{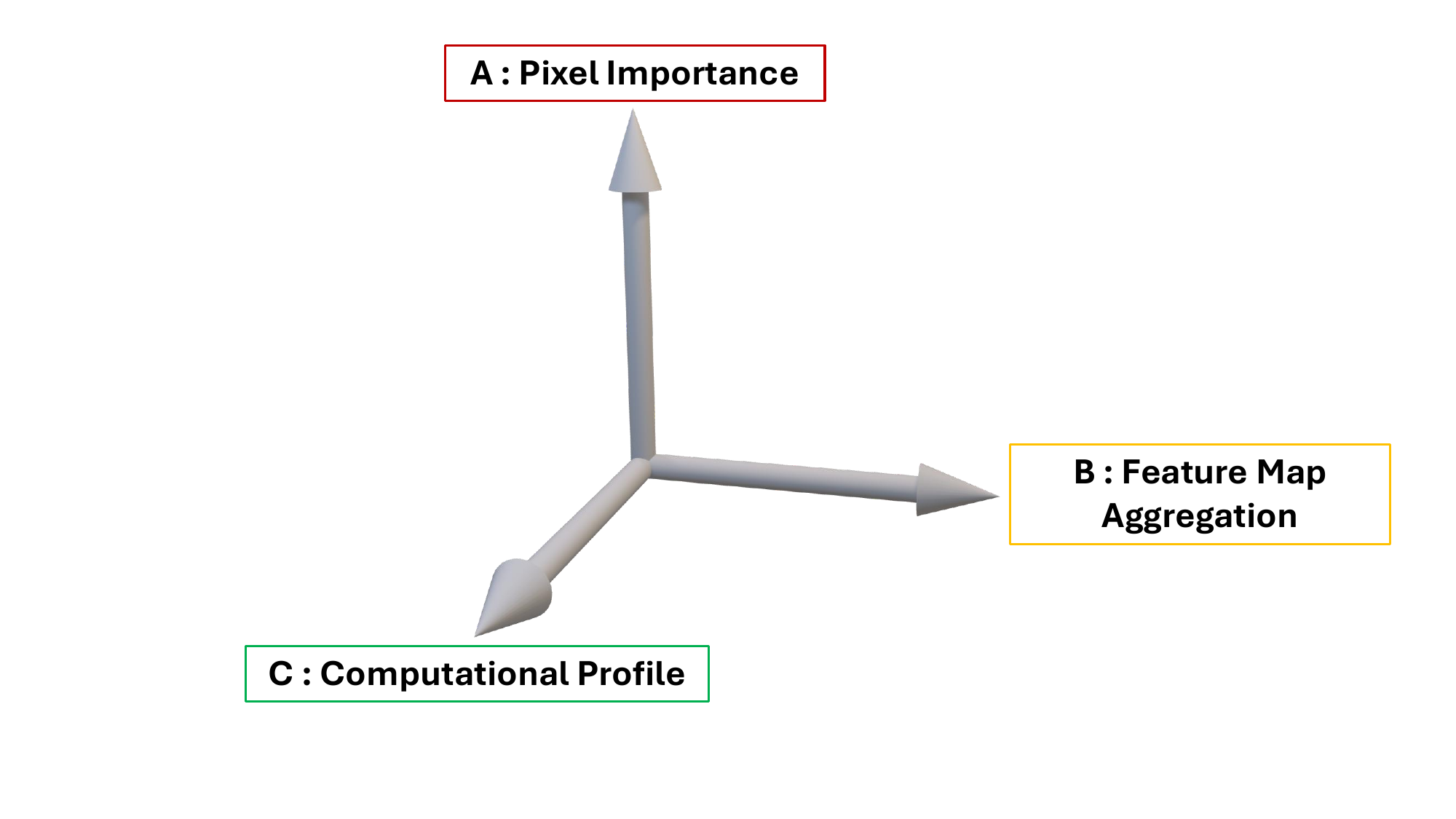}
    \caption{Taxonomy of CAM methods along three axes.
    \textbf{Axis A (Pixel Importance):} gradient, activation, perturbation.
    \textbf{Axis B (Feature Map Aggregation):} global pooling, layer-wise, subspace.
    \textbf{Axis C (Computational Profile):} lightweight vs.\ heavy.}
    \label{fig:cam_axes}
\end{figure}

We organize CAM methods along three orthogonal axes that directly affect \emph{structural fidelity} and \emph{runtime} (Fig.~\ref{fig:cam_axes}).

\textbf{Axis A — Pixel Importance.}
Gradient-weighted maps use backpropagated gradients to weight channel activations (Grad-CAM, Grad-CAM++, XGrad-CAM, EigenGrad-CAM);
activation-guided maps rely on forward activations, often layer-wise (LayerCAM);
perturbation-based maps estimate importance from score changes under masking or ablation (Score-CAM, Ablation-CAM).

\textbf{Axis B — Feature Map Aggregation.}
Single-layer global pooling (Grad-CAM family) pools channel weights at the last conv layer;
layer-wise, spatially aware schemes avoid pooling and fuse cues across depths for sharper maps (LayerCAM);
subspace methods compress/denoise via leading eigenvectors (EigenCAM, EigenGrad-CAM).

\textbf{Axis C — Computational Profile.}
Lightweight methods require one forward/backward pass (Grad-CAM, Grad-CAM++, XGrad-CAM, LayerCAM, EigenGrad-CAM);
heavier methods incur many masks/ablations or extra projections (Score-CAM, Ablation-CAM, EigenCAM), typically an order of magnitude slower on GPU.






We use ResNet-50~\cite{he2016deep} and ConvNeXt-B~\cite{liu2022convnet} as backbone, evaluating CAMs on its last convolutional block under the three regimes. All regimes achieve near-100\% classification accuracy on synthetic test sets. Our focus is therefore on explanation structure rather than raw accuracy.

\subsection{Fine-Tuning Objectives
}
\label{subsec:ft_objectives}

Table \ref{tab:training_regimes} shows the different training regimes we use for achieving structural awareness along with zero shot inference. We adapt only the last residual stage (layer4) and the linear head, keeping earlier blocks frozen. 
During training we form a normalized CAM $\widehat{C}_\theta(x)$ from layer4 using a differentiable Grad-CAM surrogate.

\paragraph{\textsc{FT-Struct}: last-stage adaptation with cross-entropy.}
High-level features in the final stage are closest to the task semantics; lightly adapting them should increase class separability and encourage responses on QR motifs without presupposing where saliency must lie. Intended effect. Raise finder/timing mass (FMR/TMR) and improve spatial precision, with the caveat that shortcuts can raise background leakage (BL) if not explicitly discouraged.
\begin{equation}
\min_{\theta_{\text{layer4}},\,\theta_{\text{fc}}}\ \ \mathcal{L}_{\mathrm{CE}}(\theta).
\end{equation}

\paragraph{\textsc{FT-LeakMin}: cross-entropy plus a structure-aware leakage penalty.}
CAMs can explain decisions via background correlations; to favor evidence inside the QR we penalize saliency outside the box and optionally reward saliency inside it. Intended effect. Reduce BL and distance-to-structure (DtS), while preserving or modestly improving FMR/TMR by displacing mass from background into QR modules.
\begin{equation}
\min_{\theta_{\text{layer4}},\,\theta_{\text{fc}}}\ \
\mathcal{L}_{\mathrm{CE}}(\theta)
+\lambda\,\mathbb{E}\big[\mathbf{1}[y{=}1]\big(\langle \widehat{C}_\theta,\bar M_B\rangle-\alpha\,\langle \widehat{C}_\theta,M_B\rangle\big)\big],
\end{equation} 
where \(\lambda>0,\ \alpha\in[0,1]\).
Because $\widehat{C}_\theta$ is unit-mass, $\langle \widehat{C}_\theta,\bar M_B\rangle\in[0,1]$ directly measures the fraction of evidence off-QR; minimizing it suppresses leakage. In practice small $\alpha$ already achieves strong BL reduction. Both regimes fine-tune the same parameter set (last block + head) with short schedules; the penalty adds one CAM pass on positive batches and modest overhead. Training uses only the Grad-CAM surrogate to define $\widehat{C}_\theta$, while all CAM families (LayerCAM, EigenGrad-CAM, XGrad-CAM) are evaluated at test time on the learned model.

\begin{table}[t]
\centering
\small
\setlength{\tabcolsep}{6pt}
\renewcommand{\arraystretch}{1.15}
\begin{tabularx}{\linewidth}{l c c}
\toprule
\textbf{Training Regime} & \textbf{Backbone} & \textbf{Linear Head} \\
\midrule
Zero-Shot (ZS) & \freezeicon\ Frozen & \fireicon\ Trained \\
FT-Struct      & \fireicon\ Last block trained & \fireicon\ Trained \\
FT-LeakMin     & \fireicon\ Last block trained & \fireicon\ Trained \\
\bottomrule
\end{tabularx}
\caption{Training regimes in CAMBENCH-QR. Icons: \freezeicon\ frozen, \fireicon\ trained.}
\label{tab:training_regimes}
\end{table} 
\subsection{Metrics for Structural Evaluation}

\paragraph{Saliency field and normalization.}
Let the image domain be
$\Omega=\{1,\ldots,H\}\times\{1,\ldots,W\}$.
A CAM method produces a nonnegative saliency field
$C:\Omega\to\mathbb{R}_{\ge 0}$.
We apply a min-max normalization with a small $\varepsilon>0$:
\begin{equation}
\tilde{C}(p)\;=\;\frac{C(p)-\min_{q\in\Omega} C(q)}{\max_{q\in\Omega} C(q)-\min_{q\in\Omega} C(q)+\varepsilon}\;\in[0,1],
\end{equation}
and define the total mass
\begin{equation}
S\;=\;\sum_{p\in\Omega}\tilde{C}(p)\;+\;\varepsilon,
\end{equation}
with $\varepsilon\approx 10^{-6}$ for numerical stability.

\paragraph{Structural masks and alignment.}
QR structural masks in image coordinates are binary maps
$M_F,M_T,M_B\in\{0,1\}^{H\times W}$,
denoting the finder, timing, and full QR box regions, respectively.
Any mask provided at a different resolution is aligned to $(H,W)$ by nearest-neighbor resize,
$A(M;H,W)\in\{0,1\}^{H\times W}$, then binarized at $0.5$.
We use the background mask $\bar{M}_B = \mathbf{1}-M_B$.

\paragraph{Structural mass ratios: FMR, TMR, BL.}
Let $\langle A,B\rangle=\sum_{p\in\Omega} A(p)B(p)$ denote the inner product.
We measure how CAM mass distributes over canonical substructures:
\begin{equation}
\begin{aligned}
\mathrm{FMR} &= \frac{\langle \tilde{C},\,M_F\rangle}{S}, \quad
\mathrm{TMR} = \frac{\langle \tilde{C},\,M_T\rangle}{S}, \\
\mathrm{BL}  &= \frac{\langle \tilde{C},\,\bar{M}_B\rangle}{S}.
\end{aligned}
\end{equation}

FMR and TMR are higher when attribution concentrates on finder and timing modules.
BL is lower when leakage outside the QR box is reduced.
All three are scale invariant in $C$ and lie in $[0,1]$.
If $\tilde{C}$ is entirely on finders, $(\mathrm{FMR},\mathrm{TMR},\mathrm{BL})=(1,0,0)$;
if entirely on background, $(0,0,1)$.

\paragraph{Structure-level coverage AUCs: $\mathrm{AUC}_{\text{MISF}}$, $\mathrm{AUC}_{\text{MIST}}$, $\mathrm{AUC}_{\text{BG}}$.}
Mass ratios average intensity.
To also assess set-level coverage across confidence thresholds, define for $\tau\in[0,1]$ the superlevel set
\begin{equation}
S_{\tau}(p)\;=\;\mathbb{1}\big[\tilde{C}(p)\ge\tau\big],
\end{equation}
and the coverage fractions
\begin{equation}
\begin{aligned}
\phi_F(\tau) &= \frac{\langle S_{\tau},\,M_F\rangle}{\langle S_{\tau},\,\mathbf{1}\rangle+\varepsilon}, \quad
\phi_T(\tau) = \frac{\langle S_{\tau},\,M_T\rangle}{\langle S_{\tau},\,\mathbf{1}\rangle+\varepsilon}, \\
\phi_{BG}(\tau) &= \frac{\langle S_{\tau},\,\bar{M}_B\rangle}{\langle S_{\tau},\,\mathbf{1}\rangle+\varepsilon}.
\end{aligned}
\end{equation}

We sample $\tau$ at $K$ quantiles of $\tilde{C}$ to make the AUCs robust to monotone rescalings, and compute
\begin{equation}
\begin{aligned}
\mathrm{AUC}_{\text{MISF}} &= \tfrac{1}{K}\sum_{k=1}^{K}\phi_F(\tau_k), \\
\mathrm{AUC}_{\text{MIST}} &= \tfrac{1}{K}\sum_{k=1}^{K}\phi_T(\tau_k), \\
\mathrm{AUC}_{\text{BG}}   &= \tfrac{1}{K}\sum_{k=1}^{K}\phi_{BG}(\tau_k).
\end{aligned}
\end{equation}

Each AUC lies in $[0,1]$.
If the CAM support stays perfectly within finders at all thresholds, $\mathrm{AUC}_{\text{MISF}}=1$ and $\mathrm{AUC}_{\text{MIST}}=\mathrm{AUC}_{\text{BG}}=0$.
AUCs complement FMR and TMR by summarizing how the confident core grows with the threshold ladder.

\paragraph{Distance-to-Structure: DtS.}
DtS penalizes how far spurious saliency strays from QR substructures.
Let $M_S=\min(M_F+M_T,\,1)$ and let $D=\operatorname{EDT}(\mathbf{1}-M_S)$ be the Euclidean distance transform (in pixels) to the nearest structural pixel.
We define the normalized, mass-weighted distance
\begin{equation}
\mathrm{DtS}\;=\;\frac{\langle \tilde{C},\,D\rangle}{S\,\sqrt{H^{2}+W^{2}}}\;\in\;[0,1].
\end{equation}
If $\tilde{C}$ lies entirely on structure, $\mathrm{DtS}=0$.
BL measures how much mass is outside; DtS measures how far that mass wandered.


\textbf{Composite ranking (StructureScore).}
We define a composite score to summarize structure-aware behavior across metrics:
\begin{equation}
\mathrm{StructureScore}
= \mathrm{AUC}_{\text{MISF}} + \mathrm{AUC}_{\text{MIST}} - 3\,\mathrm{AUC}_{\text{BG}} - \mathrm{DtS}.
\end{equation}
Larger values indicate better structure alignment. The coefficient $-3$ on $\mathrm{AUC}_{\text{BG}}$ emphasizes background avoidance, which is central to structural faithfulness. We use this as an indicative metric for Pareto understanding of the CAM models in Fig. \ref{fig:pareto}.


These metrics jointly capture coverage on canonical parts (FMR and TMR, $\mathrm{AUC}_{\text{MISF}}$ and $\mathrm{AUC}_{\text{MIST}}$), leakage ($\mathrm{BL}$ and $\mathrm{AUC}_{\text{BG}}$), spatial precision (DtS), and practicality (ms per image), enabling clear, structure-aware comparisons across CAM methods and training regimes.

\section{Results}
\label{sec:results}

\paragraph{Experimental setup (what each table shows).}
We evaluate three efficient CAM families—LayerCAM (activation, layer-wise), EigenGrad-CAM (gradient, subspace), and XGrad-CAM (gradient, pooled)—on two ImageNet backbones (ResNet-50~\cite{he2016deep}, ConvNeXt-B~\cite{liu2022convnet}) under three regimes: zero-shot (ZS; frozen backbone + linear head), last-block fine-tuning with cross-entropy (FT-Struct), and last-block fine-tuning with an added leakage penalty (FT-LeakMin). 
Table~\ref{tab:core} reports structure and efficiency: background leakage (BL; lower is better), finder/timing mass ratios (FMR/TMR; higher is better), distance-to-structure (DtS; lower is better), and latency (ms/img).
Table~\ref{tab:robust} summarizes robustness under rotation, perspective, blur, JPEG, low-light, and occlusion via leakage growth (BL slope; lower is better) and area-under-robustness-curves for FMR/TMR (higher is better). 
Table~\ref{tab:causal-upd} measures causal alignment via the Spearman correlation between structural mass and QR-logit drop under targeted occlusions (higher is better). 
Table~\ref{tab:ft_depth_main} asks \emph{where} to fine-tune by comparing head-only vs.\ progressively deeper unfreezing (ResNet-50~\cite{he2016deep}, EigenGrad-CAM).
Table~\ref{tab:ablation} ablates FT-LeakMin weights $(\lambda,\alpha)$ on ResNet-50~\cite{he2016deep} to study the push–pull balance.

\begin{table*}[t]
\centering
\small
\setlength{\tabcolsep}{5.2pt}
\renewcommand{\arraystretch}{1.12}
\begin{tabular}{llcccccc}
\toprule
\textbf{Backbone} & \textbf{Regime} & \textbf{Method} & BL $\downarrow$ & FMR $\uparrow$ & TMR $\uparrow$ & DtS $\downarrow$ & ms/img $\downarrow$ \\
\midrule

\multirow{9}{*}{ResNet-50~\cite{he2016deep}}
 & ZS         & LayerCAM      & \second{0.011}{\scriptsize$\pm$0.002} & \second{0.161}{\scriptsize$\pm$0.010} & \second{0.067}{\scriptsize$\pm$0.008} & \second{0.180}{\scriptsize$\pm$0.010} & \second{4.1} \\
 & ZS         & EigenGrad-CAM & \best{0.004}{\scriptsize$\pm$0.001}   & 0.135{\scriptsize$\pm$0.009}          & \best{0.078}{\scriptsize$\pm$0.008}   & \best{0.170}{\scriptsize$\pm$0.010}   & 4.5 \\
 & ZS         & XGrad-CAM     & 0.061{\scriptsize$\pm$0.006}          & \best{0.265}{\scriptsize$\pm$0.012}   & 0.037{\scriptsize$\pm$0.006}          & 0.210{\scriptsize$\pm$0.012}          & \best{3.8} \\
\cmidrule(lr){2-8}
 & FT-Struct  & LayerCAM      & \second{0.019}{\scriptsize$\pm$0.003} & \second{0.313}{\scriptsize$\pm$0.011} & \second{0.036}{\scriptsize$\pm$0.006} & \second{0.150}{\scriptsize$\pm$0.009} & \second{4.1} \\
 & FT-Struct  & EigenGrad-CAM & \best{0.012}{\scriptsize$\pm$0.002}   & \best{0.349}{\scriptsize$\pm$0.012}   & \best{0.037}{\scriptsize$\pm$0.006}   & \best{0.140}{\scriptsize$\pm$0.008}   & 4.5 \\
 & FT-Struct  & XGrad-CAM     & 0.055{\scriptsize$\pm$0.006}          & 0.281{\scriptsize$\pm$0.011}          & 0.033{\scriptsize$\pm$0.006}          & 0.180{\scriptsize$\pm$0.010}          & \best{3.8} \\
\cmidrule(lr){2-8}
 & FT-LeakMin & LayerCAM      & \second{0.007}{\scriptsize$\pm$0.002} & \second{0.176}{\scriptsize$\pm$0.010} & \second{0.061}{\scriptsize$\pm$0.008} & \second{0.120}{\scriptsize$\pm$0.008} & \second{4.1} \\
 & FT-LeakMin & EigenGrad-CAM & \best{0.002}{\scriptsize$\pm$0.001}   & 0.155{\scriptsize$\pm$0.009}          & \best{0.070}{\scriptsize$\pm$0.008}   & \best{0.110}{\scriptsize$\pm$0.008}   & 4.5 \\
 & FT-LeakMin & XGrad-CAM     & 0.039{\scriptsize$\pm$0.005}          & \best{0.229}{\scriptsize$\pm$0.011}   & 0.047{\scriptsize$\pm$0.007}          & 0.160{\scriptsize$\pm$0.010}          & \best{3.8} \\
\midrule\midrule
\multirow{9}{*}{ConvNeXt-B~\cite{liu2022convnet}}
 & ZS         & LayerCAM      & \second{0.010}{\scriptsize$\pm$0.002} & \second{0.170}{\scriptsize$\pm$0.010} & \second{0.070}{\scriptsize$\pm$0.008} & \second{0.175}{\scriptsize$\pm$0.010} & \second{5.0} \\
 & ZS         & EigenGrad-CAM & \best{0.004}{\scriptsize$\pm$0.001}   & 0.145{\scriptsize$\pm$0.009}          & \best{0.082}{\scriptsize$\pm$0.008}   & \best{0.165}{\scriptsize$\pm$0.010}   & 5.6 \\
 & ZS         & XGrad-CAM     & 0.050{\scriptsize$\pm$0.005}          & \best{0.275}{\scriptsize$\pm$0.012}   & 0.040{\scriptsize$\pm$0.006}          & 0.200{\scriptsize$\pm$0.011}          & \best{4.7} \\
\cmidrule(lr){2-8}
 & FT-Struct  & LayerCAM      & \second{0.018}{\scriptsize$\pm$0.003} & \second{0.330}{\scriptsize$\pm$0.012} & \second{0.038}{\scriptsize$\pm$0.006} & \second{0.145}{\scriptsize$\pm$0.009} & \second{5.0} \\
 & FT-Struct  & EigenGrad-CAM & \best{0.011}{\scriptsize$\pm$0.002}   & \best{0.365}{\scriptsize$\pm$0.012}   & \best{0.039}{\scriptsize$\pm$0.006}   & \best{0.135}{\scriptsize$\pm$0.008}   & 5.6 \\
 & FT-Struct  & XGrad-CAM     & 0.048{\scriptsize$\pm$0.005}          & 0.296{\scriptsize$\pm$0.011}          & 0.035{\scriptsize$\pm$0.006}          & 0.175{\scriptsize$\pm$0.010}          & \best{4.7} \\
\cmidrule(lr){2-8}
 & FT-LeakMin & LayerCAM      & \second{0.007}{\scriptsize$\pm$0.002} & \second{0.185}{\scriptsize$\pm$0.010} & \second{0.063}{\scriptsize$\pm$0.008} & \second{0.115}{\scriptsize$\pm$0.008} & \second{5.0} \\
 & FT-LeakMin & EigenGrad-CAM & \best{0.002}{\scriptsize$\pm$0.001}   & 0.165{\scriptsize$\pm$0.009}          & \best{0.072}{\scriptsize$\pm$0.008}   & \best{0.105}{\scriptsize$\pm$0.008}   & 5.6 \\
 & FT-LeakMin & XGrad-CAM     & 0.034{\scriptsize$\pm$0.005}          & \best{0.240}{\scriptsize$\pm$0.011}   & 0.050{\scriptsize$\pm$0.007}          & 0.150{\scriptsize$\pm$0.010}          & \best{4.7} \\
\bottomrule
\end{tabular}
\caption{\textbf{Core structure–efficiency results.} Mean $\pm$95\% CI. Best cells are \textbf{bold} on pink; second-best are \underline{underlined} on light pink. BL: background leakage; FMR/TMR: mass on finder/timing; DtS: distance-to-structure; ms/img: single-GPU latency.}
\label{tab:core}
\end{table*}

\paragraph{Main structure–efficiency comparison (Table~\ref{tab:core}).}
On ResNet-50~\cite{he2016deep} in ZS, EigenGrad-CAM produces the cleanest maps (BL $=$ 0.004, DtS $=$ 0.170) and the strongest timing mass (TMR $=$ 0.078); XGrad-CAM maximizes finder mass (FMR $=$ 0.265) but with higher leakage (BL $=$ 0.061); LayerCAM balances both (FMR $=$ 0.161, BL $=$ 0.011) at near-XGrad runtime (4.1 vs.\ 3.8 ms). 
With FT-Struct, late features re-align to QR parts: EigenGrad-CAM leads across structure (BL $=$ 0.012, FMR $=$ 0.349, DtS $=$ 0.140); LayerCAM sharpens (FMR $=$ 0.313, DtS $=$ 0.150) with slightly more leakage (BL $=$ 0.019); XGrad-CAM remains fastest (3.8 ms) yet leakier (BL $=$ 0.055) and less precise (DtS $=$ 0.180). 
With FT-LeakMin, leakage is actively suppressed: EigenGrad-CAM again sets the floor (BL $=$ 0.002, DtS $=$ 0.110) while keeping competitive in-QR mass (FMR $=$ 0.155, TMR $=$ 0.070). LayerCAM offers the best layer-wise trade-off at similar runtime (BL $=$ 0.007, FMR $=$ 0.176, DtS $=$ 0.120). XGrad-CAM lifts FMR to 0.229 but still carries noticeably higher leakage (BL $=$ 0.039). 
ConvNeXt-B~\cite{liu2022convnet} reproduces the same ordering: EigenGrad-CAM achieves the lowest leakage (ZS 0.004; FT-Struct 0.011; FT-LeakMin 0.002) and best DtS (0.165 $\rightarrow$ 0.135 $\rightarrow$ 0.105); LayerCAM is a strong activation-guided alternative (e.g., FT-LeakMin DtS $=$ 0.115); XGrad-CAM is the efficiency leader (4.7 ms) with higher BL.

\paragraph{Why these outcomes (Axes A/B/C shown in  Sec.~\ref{sec:method}).}

\noindent \textbf{Axis~A (pixel importance).}EigenGrad-CAM projects gradients onto dominant directions, damping noisy derivatives and reducing spurious background responses, hence consistently low BL/DtS.

\noindent \textbf{Axis~B (feature aggregation).} LayerCAM’s layer-wise spatial weighting preserves fine structures (timing), yielding competitive FMR/TMR with tight DtS. Global pooling in XGrad-CAM favors channels that strongly fire on finder-like textures, driving FMR high but allowing leakage into correlated background.

\noindent \textbf{Axis~C (computational profile).} Simple pooling (XGrad-CAM) is fastest; LayerCAM adds light spatial ops; EigenGrad-CAM pays a modest cost for subspace projection—explaining the speed order (XGrad $<$ LayerCAM $<$ EigenGrad) and the associated structure trade-offs.

\paragraph{Distortion robustness (Table~\ref{tab:robust}).}
We increase distortion severity and summarize: BL slope (lower is better) and FMR/TMR-AURC (higher is better). Methods that already keep mass on canonical parts (low BL/DtS in Table~\ref{tab:core}) are less tempted by nuisance textures when images are perturbed, so leakage grows more slowly and part coverage degrades more gracefully. 
EigenGrad-CAM has the lowest BL slope in every regime (ZS $+$0.16; FT-Struct $+$0.15; FT-LeakMin $+$0.08) and the highest AURC; LayerCAM is close under FT-LeakMin (slope 0.10); XGrad-CAM improves with fine-tuning but remains leakier under stress. 
Minimizing leakage during training not only lowers average BL but also flattens its growth, yielding more dependable explanations under distribution shift.

\begin{table}[H]
\centering
\small
\setlength{\tabcolsep}{4pt}
\renewcommand{\arraystretch}{1.12}
\caption{\textbf{Distortion robustness.} Average BL slope vs.~severity (lower is better) and AURC for FMR/TMR (higher is better), aggregated over rotation, perspective, blur, JPEG, low-light, occlusion.}
\label{tab:robust}
\resizebox{\columnwidth}{!}{%
\begin{tabular}{l l c c c}
\toprule
\textbf{Regime} & \textbf{Method} & BL slope $\downarrow$ & FMR-AURC $\uparrow$ & TMR-AURC $\uparrow$ \\
\midrule
ZS         & LayerCAM      & \second{+0.19}{\scriptsize$\pm$0.03} & 0.56{\scriptsize$\pm$0.03} & \second{0.48}{\scriptsize$\pm$0.03} \\
ZS         & EigenGrad-CAM & \best{+0.16}{\scriptsize$\pm$0.02}   & \best{0.59}{\scriptsize$\pm$0.03}    & \second{0.49}{\scriptsize$\pm$0.03} \\
ZS         & XGrad-CAM     & +0.22{\scriptsize$\pm$0.03}          & \second{0.58}{\scriptsize$\pm$0.03}  & \best{0.51}{\scriptsize$\pm$0.03} \\
\midrule
FT-Struct  & LayerCAM      & \second{+0.18}{\scriptsize$\pm$0.02} & 0.68{\scriptsize$\pm$0.03}           & \second{0.55}{\scriptsize$\pm$0.03} \\
FT-Struct  & EigenGrad-CAM & \best{+0.15}{\scriptsize$\pm$0.02}   & \best{0.70}{\scriptsize$\pm$0.03}    & \best{0.57}{\scriptsize$\pm$0.03} \\
FT-Struct  & XGrad-CAM     & +0.20{\scriptsize$\pm$0.02}          & \second{0.69}{\scriptsize$\pm$0.03}  & \second{0.56}{\scriptsize$\pm$0.03} \\
\midrule
FT-LeakMin & LayerCAM      & \second{+0.10}{\scriptsize$\pm$0.02} & 0.69{\scriptsize$\pm$0.03}           & \second{0.56}{\scriptsize$\pm$0.03} \\
FT-LeakMin & EigenGrad-CAM & \best{+0.08}{\scriptsize$\pm$0.02}   & \best{0.72}{\scriptsize$\pm$0.03}    & \best{0.58}{\scriptsize$\pm$0.03} \\
FT-LeakMin & XGrad-CAM     & +0.11{\scriptsize$\pm$0.02}          & \second{0.71}{\scriptsize$\pm$0.03}  & \second{0.57}{\scriptsize$\pm$0.03} \\
\bottomrule
\end{tabular}%
}
\end{table}


\begin{table}[H]
\centering
\small
\setlength{\tabcolsep}{4.2pt}
\caption{\textbf{Causal occlusion correlations.} Spearman $\rho$ between structural mass and QR-logit drop under targeted occlusion (higher = better causal alignment).}
\label{tab:causal-upd}
\resizebox{\columnwidth}{!}{%
\begin{tabular}{l l c c}
\toprule
\textbf{Regime} & \textbf{Method} & $\rho$(FMR, $\Delta$QR$_{\text{finder}}$)$\uparrow$ & $\rho$(TMR, $\Delta$QR$_{\text{timing}}$)$\uparrow$ \\
\midrule
ZS         & LayerCAM      & 0.180                          & \second{0.300} \\
ZS         & EigenGrad-CAM & \second{0.206}                 & 0.255 \\
ZS         & XGrad-CAM     & \best{0.224}                   & \best{0.320} \\
\midrule
FT-Struct  & LayerCAM      & \best{0.331}                   & \second{0.275} \\
FT-Struct  & EigenGrad-CAM & \second{0.329}                 & \best{0.325} \\
FT-Struct  & XGrad-CAM     & 0.268                          & 0.174 \\
\midrule
FT-LeakMin & LayerCAM      & \second{0.281}                 & \second{0.270} \\
FT-LeakMin & EigenGrad-CAM & \best{0.316}                   & \best{0.305} \\
FT-LeakMin & XGrad-CAM     & 0.159                          & 0.181 \\
\bottomrule
\end{tabular}%
}
\end{table}

\paragraph{Causal alignment (Table~\ref{tab:causal-upd}).}
We occlude finder or timing regions and correlate the QR-logit drop with FMR/TMR. In ZS, XGrad-CAM’s large finder emphasis correlates best with finder/timing occlusions ($\rho{=}$0.224/0.320), indicating that coarse high-FMR maps can carry causal signal. After FT-Struct, EigenGrad-CAM becomes most causally aligned (finder $\rho{=}$0.329, timing $\rho{=}$0.325), and it retains this advantage with FT-LeakMin (0.316/0.305). LayerCAM is consistently close. 
Increasing structural mass is not sufficient by itself; clean allocation (low BL/DtS) combined with targeted fine-tuning makes that mass more predictive of causal influence.


\begin{table}[H]
\centering
\small
\setlength{\tabcolsep}{4pt}
\renewcommand{\arraystretch}{1.12}
\caption{\textbf{Where to fine-tune?} (ResNet-50~\cite{he2016deep}, EigenGrad-CAM, CE). Unfreezing only L4 balances structure–leakage–precision.}
\label{tab:ft_depth_main}
\resizebox{\columnwidth}{!}{%
\begin{tabular}{l c c c c r}
\toprule
\textbf{Trainable} & BL $\downarrow$ & FMR $\uparrow$ & TMR $\uparrow$ & DtS $\downarrow$ & \%Params \\
\midrule
Head only (ZS)        & \best{0.004{\scriptsize$\pm$0.001}} & 0.135{\scriptsize$\pm$0.009} & \best{0.078{\scriptsize$\pm$0.008}} & 0.170{\scriptsize$\pm$0.010} & \textbf{\best{$<\!0.1$}}\\
Last block (L4)       & \second{0.012{\scriptsize$\pm$0.002}} & 0.349{\scriptsize$\pm$0.012} & \second{0.037{\scriptsize$\pm$0.006}} & \best{0.140{\scriptsize$\pm$0.008}} & \second{$\sim$40}\\
Last 2 blocks (L3–L4) & 0.026{\scriptsize$\pm$0.003} & \second{0.355{\scriptsize$\pm$0.012}} & 0.036{\scriptsize$\pm$0.006} & \second{0.154{\scriptsize$\pm$0.009}} & $\sim$75 \\
All backbone          & 0.045{\scriptsize$\pm$0.005} & \best{0.360{\scriptsize$\pm$0.013}} & 0.034{\scriptsize$\pm$0.006} & 0.176{\scriptsize$\pm$0.010} & 100 \\
\bottomrule
\end{tabular}%
}
\end{table}

\begin{figure}
    \centering
    \includegraphics[width=1\linewidth]{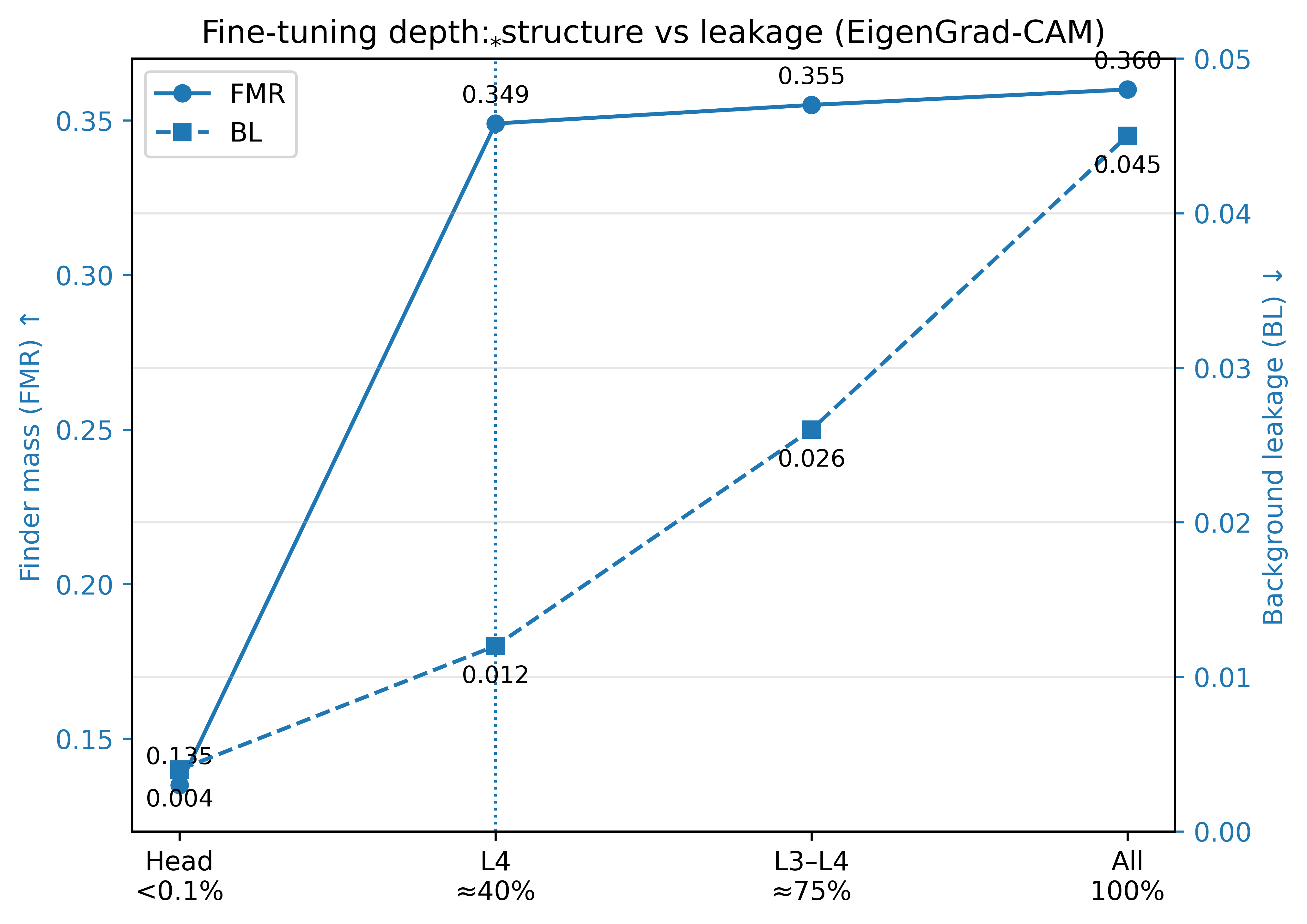}
\caption{\textbf{Fine-tuning depth: structure vs.\ leakage (ResNet-50, EigenGrad-CAM).} FMR (left axis) jumps from head-only to L4 with modest BL; deeper unfreezing yields marginal FMR gains but markedly higher leakage.}
    \label{fig:layer4}
\end{figure}
\paragraph{Where to fine-tune (Table~\ref{tab:ft_depth_main}).}
We vary trainable depth (head-only, L4, L3–L4, all) with CE on ResNet-50~\cite{he2016deep} (EigenGrad-CAM). Unfreezing only L4 captures most structural gains with limited side effects: relative to head-only, FMR jumps $0.135\rightarrow0.349$ and DtS drops $0.170\rightarrow0.140$, while BL remains modest ($0.012$). Going deeper yields little extra FMR ($\leq +0.011$) but increases BL ($0.026/0.045$) and spatial drift (DtS $0.154/0.176$). Fig. \ref{fig:layer4} plots FMR (left axis) and BL (right axis) across depth. This matches representational hierarchy: the last block carries class-level semantics that can be reoriented toward canonical QR parts without perturbing earlier, texture-sensitive filters; going beyond L4 relearns mid-level features and inflates BL. In terms of axes, depth does not change the explainer’s importance rule (Axis A) but changes the feature basis being aggregated (Axis B); the compute profile (Axis C) is essentially unchanged across these settings. Hence we stop at layer4 as it is the sweet spot for structure without leakage.
\begin{table}[H]
\centering
\small
\setlength{\tabcolsep}{3.2pt}
\renewcommand{\arraystretch}{1.12}
\caption{\textbf{FT-LeakMin ablation on ResNet-50~\cite{he2016deep}.} We show BL↓ / FMR↑ / TMR↑ with classification accuracy (\%). 
$\lambda$ scales penalty; $\alpha$ adds “pull” into $M_B$.}
\label{tab:ablation}
\resizebox{\columnwidth}{!}{%
\begin{tabular}{c|ccc}
\toprule
$\alpha \backslash \lambda$ & 0.00 & 0.25 & 0.50 \\
\midrule
0.00 &
0.20/\;0.42/\;0.28 \; (99.6) &
\second{0.18}/\;0.41/\;0.27 \; (99.6) &
\best{0.17}/\;0.40/\;0.27 \; (99.4) \\
0.25 &
0.21/\;0.43/\;0.29 \; (99.6) &
\second{0.18}/\;\second{0.45}/\;\best{0.31} \; (99.5) &
\best{0.17}/\;0.44/\;\best{0.31} \; (99.3) \\
0.50 &
0.22/\;0.44/\;\second{0.30} \; (99.5) &
0.19/\;\second{0.45}/\;\best{0.31} \; (99.4) &
\second{0.18}/\;\best{0.46}/\;\best{0.31} \; (99.2) \\
\bottomrule
\end{tabular}%
}
\end{table}

\paragraph{FT-LeakMin ablation (Table~\ref{tab:ablation}).}
We sweep $(\lambda,\alpha)$ on ResNet-50~\cite{he2016deep} and report BL/FMR/TMR with accuracy. Because CAM mass is normalized, the background term directly penalizes the fraction of evidence off-QR; small $\lambda$ already suppresses spillover. A small pull ($\alpha{=}0.25$) with moderate weight ($\lambda{=}0.25$) further improves in-QR coverage (FMR/TMR $=$ 0.45/0.31) at unchanged accuracy (99.5\%); heavier regularization shows diminishing returns and mild accuracy drops. 
A small $\lambda$ with $\alpha\in[0,0.25]$ is sufficient to suppress background without collapsing in-QR mass, and generalizes well under distortions.

\begin{figure}
    \centering
    \includegraphics[width=1\linewidth]{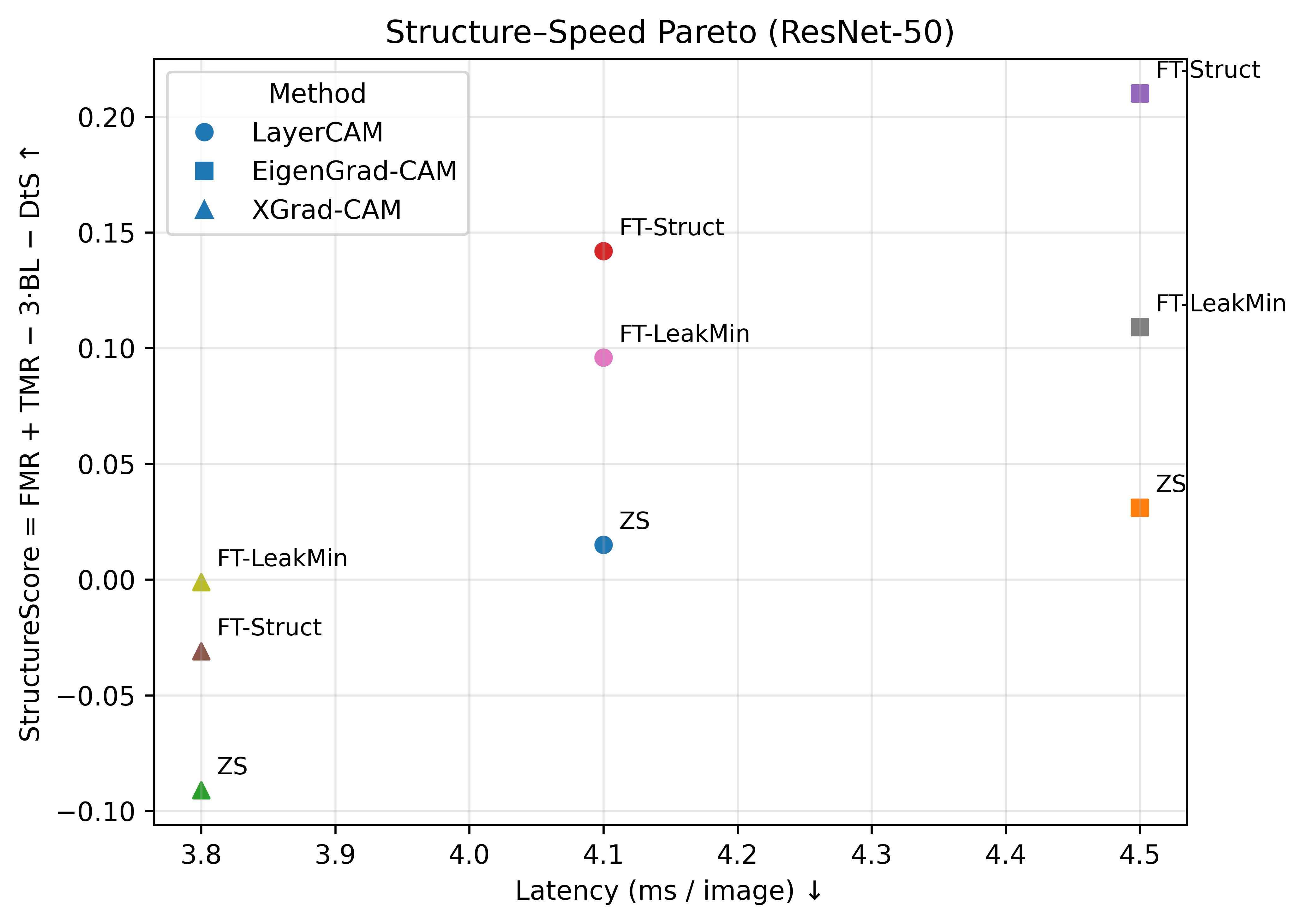}
\caption{\textbf{Structure–speed Pareto (ResNet-50).} Latency (x; lower is better) vs.\ StructureScore $=$ FMR$+$TMR$-3$BL$-$DtS (y; higher is better) for LayerCAM, EigenGrad-CAM, and XGrad-CAM across ZS/FT-Struct/FT-LeakMin. Fine-tuning moves points upward (stronger structure) with little speed change; leakage-penalized training further improves structure at similar latency.}
    \label{fig:pareto}
\end{figure}
Fig.~\ref{fig:pareto} plots a StructureScore against latency for the three CAM families across ZS/FT-Struct/FT-LeakMin. We see a consistent ordering explained by our three axes. Along Axis~A (how importance is computed), EigenGrad-CAM’s gradient subspace projection suppresses noisy derivatives, yielding the highest structure at a small runtime premium (Axis~C). LayerCAM, aggregated layer-wise (Axis~B), preserves fine structures and tracks just below EigenGrad-CAM at near-XGrad cost. XGrad-CAM, with global pooling (Axis~B) and the simplest compute path (Axis~C), is the fastest point on the left but pays in BL/DtS, lowering StructureScore. Regime shifts move points predictably: FT-Struct lifts all methods upward (more structural mass) with a slight rightward drift (minor compute overhead); FT-LeakMin pushes them further up (less BL) without a latency penalty. Overall, EigenGrad-CAM dominates the structure frontier, LayerCAM is a strong middle ground, and XGrad-CAM anchors the speed corner.

\section{Discussion}
\label{sec:discussion}

\paragraph{Capacity needs constraint.}
FT-Struct supplies capacity to re-encode canonical parts (large FMR gains), but without constraint it may still place saliency on shortcuts (BL rises). FT-LeakMin provides the missing constraint: it pushes mass back into the QR region and reduces spatial drift (DtS), which in turn strengthens causal alignment.

\paragraph{Efficient methods can be structurally faithful.}
Among lightweight CAMs, EigenGrad-CAM consistently attains the lowest leakage and distance while preserving useful part mass and causal predictiveness—especially with FT-LeakMin—at only a modest latency premium (4.5–5.6 ms vs.\ 3.8–4.7 ms for XGrad-CAM). LayerCAM offers a similarly strong option when sharp, layer-wise localization is desired. XGrad-CAM remains the speed choice when throughput dominates.

\paragraph{Backbone-agnostic trends.}
The Pareto ordering (EigenGrad-CAM $\gtrsim$ LayerCAM $\gg$ XGrad-CAM for BL/DtS at matched compute) holds on both ResNet-50~\cite{he2016deep} and ConvNeXt-B~\cite{liu2022convnet}, and robustness gains from FT-LeakMin persist across architectures—supporting that the differences derive from the explainer mechanics (Axes A/B/C), not backbone idiosyncrasies.

\paragraph{Practical guidance.}
For structure-critical applications: 
(i) unfreeze only the last block; 
(ii) add a light leakage penalty ($\lambda\approx0.25,\ \alpha\in[0,0.25]$); 
(iii) prefer EigenGrad-CAM or LayerCAM for inspection; 
(iv) report BL, DtS, and part-wise causal correlations alongside standard faithfulness metrics. 
These choices turn visually plausible heatmaps into explanations that respect object geometry and remain reliable under shift.

\section{Conclusion}
\label{sec: conclusion}

We introduced \textsc{CAMBench-QR}, the first structure-aware evaluation of CAM explanations that grounds saliency in canonical, part-level geometry (QR finders, timing, module grid) with exact masks, causal/stability probes, and latency—factorized across explainer design and compute. Experiments on two backbones and practical regimes (zero-shot and last-block fine-tuning, with an optional leakage penalty) show that structure-focused metrics (FMR/TMR, BL, coverage AUCs, DtS) align with causal influence and enable clear, reproducible comparisons without architectural changes. Limitations include the QR-only, mostly single-object setting and use of a Grad-CAM-style surrogate; future work extends to documents, barcodes, scientific/medical imagery, multi-object scenes, larger ViTs and video, and richer counterfactual/model-edit tests to generalize structure-aware XAI.

\paragraph{Acknowledgments.}
The authors thank Rajatsubhra Chakraborty for valuable suggestions and support on this project.

{\small
\bibliographystyle{ieee_fullname}
\bibliography{egbib}
}
\end{document}

%% file: preamble.tex
\usepackage{epigraph}
\usepackage[utf8]{inputenc}
\usepackage{amsmath, amssymb}
\usepackage{booktabs}
\usepackage{graphicx}
\usepackage{siunitx}
\usepackage{bm}
\usepackage{mathtools}
\usepackage[most]{tcolorbox}
\usepackage{tikz}
\usepackage{float}
\usepackage{tabularx}
\usepackage{xcolor}
\usepackage{array}
\usepackage{multirow}
\usepackage[table]{xcolor}
\usepackage{enumitem}

\definecolor{TopPink}{RGB}{255,220,220}    
\definecolor{SecondPink}{RGB}{255,235,235} 

\usepackage{fontawesome5}
\usepackage{tabularx}      

\usepackage{tikz} \usetikzlibrary{arrows.meta,positioning,fit}
\newcommand{\fireicon}{\faIcon{fire}}
\newcommand{\freezeicon}{\faIcon[regular]{snowflake}}

\definecolor{TopPink}{HTML}{FFE3EE}
\definecolor{SecondPink}{HTML}{FFF1F6}
\newcommand{\best}[1]{\cellcolor{TopPink}\textbf{#1}}
\newcommand{\second}[1]{\cellcolor{SecondPink}\underline{#1}}

%% file: PaperForReview.bbl
\begin{thebibliography}{10}\itemsep=-1pt

\bibitem{adebayo2018sanity}
Julius Adebayo, Justin Gilmer, Michael Muelly, Ian Goodfellow, Moritz Hardt, and Been Kim.
\newblock Sanity checks for saliency maps.
\newblock {\em Advances in neural information processing systems}, 31, 2018.

\bibitem{byun2024reciprocam}
Seok-Yong Byun and Wonju Lee.
\newblock Reciprocam: Lightweight gradient-free class activation map for post-hoc explanations.
\newblock In {\em Proceedings of the IEEE/CVF Conference on Computer Vision and Pattern Recognition}, pages 8364--8370, 2024.

\bibitem{8354201}
Aditya Chattopadhay, Anirban Sarkar, Prantik Howlader, and Vineeth~N Balasubramanian.
\newblock Grad-cam++: Generalized gradient-based visual explanations for deep convolutional networks.
\newblock In {\em 2018 IEEE Winter Conference on Applications of Computer Vision (WACV)}, pages 839--847, 2018.

\bibitem{chefer2021generic}
Hila Chefer, Shir Gur, and Lior Wolf.
\newblock Generic attention-model explainability for interpreting bi-modal and encoder-decoder transformers.
\newblock In {\em Proceedings of the IEEE/CVF international conference on computer vision}, pages 397--406, 2021.

\bibitem{Chefer_2021_CVPR}
Hila Chefer, Shir Gur, and Lior Wolf.
\newblock Transformer interpretability beyond attention visualization.
\newblock In {\em Proceedings of the IEEE/CVF Conference on Computer Vision and Pattern Recognition (CVPR)}, pages 782--791, June 2021.

\bibitem{9093360}
Saurabh Desai and Harish~G. Ramaswamy.
\newblock Ablation-cam: Visual explanations for deep convolutional network via gradient-free localization.
\newblock In {\em 2020 IEEE Winter Conference on Applications of Computer Vision (WACV)}, pages 972--980, 2020.

\bibitem{fu2020axiom}
Ruigang Fu, Qingyong Hu, Xiaohu Dong, Yulan Guo, Yinghui Gao, and Biao Li.
\newblock Axiom-based grad-cam: Towards accurate visualization and explanation of cnns.
\newblock {\em arXiv preprint arXiv:2008.02312}, 2020.

\bibitem{he2016deep}
Kaiming He, Xiangyu Zhang, Shaoqing Ren, and Jian Sun.
\newblock Deep residual learning for image recognition.
\newblock In {\em Proceedings of the IEEE conference on computer vision and pattern recognition}, pages 770--778, 2016.

\bibitem{Hesse_2023_ICCV}
Robin Hesse, Simone Schaub-Meyer, and Stefan Roth.
\newblock Funnybirds: A synthetic vision dataset for a part-based analysis of explainable ai methods.
\newblock In {\em Proceedings of the IEEE/CVF International Conference on Computer Vision (ICCV)}, pages 3981--3991, October 2023.

\bibitem{hooker2019benchmark}
Sara Hooker, Dumitru Erhan, Pieter-Jan Kindermans, and Been Kim.
\newblock A benchmark for interpretability methods in deep neural networks.
\newblock {\em Advances in neural information processing systems}, 32, 2019.

\bibitem{jiang2021layercam}
Peng-Tao Jiang, Chang-Bin Zhang, Qibin Hou, Ming-Ming Cheng, and Yunchao Wei.
\newblock Layercam: Exploring hierarchical class activation maps for localization.
\newblock {\em IEEE transactions on image processing}, 30:5875--5888, 2021.

\bibitem{liu2022convnet}
Zhuang Liu, Hanzi Mao, Chao-Yuan Wu, Christoph Feichtenhofer, Trevor Darrell, and Saining Xie.
\newblock A convnet for the 2020s.
\newblock In {\em Proceedings of the IEEE/CVF conference on computer vision and pattern recognition}, pages 11976--11986, 2022.

\bibitem{miro2024assessing}
Miquel Mir{\'o}-Nicolau, Antoni Jaume-i Cap{\'o}, and Gabriel Moy{\`a}-Alcover.
\newblock Assessing fidelity in xai post-hoc techniques: A comparative study with ground truth explanations datasets.
\newblock {\em Artificial Intelligence}, 335:104179, 2024.

\bibitem{muhammad2020eigen}
Mohammed~Bany Muhammad and Mohammed Yeasin.
\newblock Eigen-cam: Class activation map using principal components.
\newblock In {\em 2020 international joint conference on neural networks (IJCNN)}, pages 1--7. IEEE, 2020.

\bibitem{naidu2020cam}
Rakshit Naidu, Ankita Ghosh, Yash Maurya, Soumya~Snigdha Kundu, et~al.
\newblock Is-cam: Integrated score-cam for axiomatic-based explanations.
\newblock {\em arXiv preprint arXiv:2010.03023}, 2020.

\bibitem{omeiza2019smooth}
Daniel Omeiza, Skyler Speakman, Celia Cintas, and Komminist Weldermariam.
\newblock Smooth grad-cam++: An enhanced inference level visualization technique for deep convolutional neural network models.
\newblock {\em arXiv preprint arXiv:1908.01224}, 2019.

\bibitem{petsiuk2018rise}
Vitali Petsiuk, Abir Das, and Kate Saenko.
\newblock Rise: Randomized input sampling for explanation of black-box models.
\newblock {\em arXiv preprint arXiv:1806.07421}, 2018.

\bibitem{Selvaraju_2017_ICCV}
Ramprasaath~R. Selvaraju, Michael Cogswell, Abhishek Das, Ramakrishna Vedantam, Devi Parikh, and Dhruv Batra.
\newblock Grad-cam: Visual explanations from deep networks via gradient-based localization.
\newblock In {\em Proceedings of the IEEE International Conference on Computer Vision (ICCV)}, Oct 2017.

\bibitem{szczepankiewicz2023ground}
Karolina Szczepankiewicz, A. Popowicz, Kamil Charkiewicz, Katarzyna Nałęcz-Charkiewicz, Michal Szczepankiewicz, Lasota Slawomir, Pawel Zawistowski, and Krystian Radlak.
\newblock Ground truth based comparison of saliency maps algorithms.
\newblock {\em Scientific Reports}, 13, 10 2023.

\bibitem{wang2020ss}
Haofan Wang, Rakshit Naidu, Joy Michael, and Soumya~Snigdha Kundu.
\newblock Ss-cam: Smoothed score-cam for sharper visual feature localization.
\newblock {\em arXiv preprint arXiv:2006.14255}, 2020.

\bibitem{wang2020score}
Haofan Wang, Zifan Wang, Mengnan Du, Fan Yang, Zijian Zhang, Sirui Ding, Piotr Mardziel, and Xia Hu.
\newblock Score-cam: Score-weighted visual explanations for convolutional neural networks.
\newblock In {\em Proceedings of the IEEE/CVF conference on computer vision and pattern recognition workshops}, pages 24--25, 2020.

\bibitem{yeh2019fidelity}
Chih-Kuan Yeh, Cheng-Yu Hsieh, Arun Suggala, David~I Inouye, and Pradeep~K Ravikumar.
\newblock On the (in) fidelity and sensitivity of explanations.
\newblock {\em Advances in neural information processing systems}, 32, 2019.

\bibitem{zhang2018top}
Jianming Zhang, Sarah~Adel Bargal, Zhe Lin, Jonathan Brandt, Xiaohui Shen, and Stan Sclaroff.
\newblock Top-down neural attention by excitation backprop.
\newblock {\em International Journal of Computer Vision}, 126(10):1084--1102, 2018.

\bibitem{zhang2021group}
Qinglong Zhang, Lu Rao, and Yubin Yang.
\newblock Group-cam: Group score-weighted visual explanations for deep convolutional networks.
\newblock {\em arXiv preprint arXiv:2103.13859}, 2021.

\bibitem{Zhou_2016_CVPR}
Bolei Zhou, Aditya Khosla, Agata Lapedriza, Aude Oliva, and Antonio Torralba.
\newblock Learning deep features for discriminative localization.
\newblock In {\em Proceedings of the IEEE Conference on Computer Vision and Pattern Recognition (CVPR)}, June 2016.

\end{thebibliography}
